\documentclass[runningheads]{llncs}
\usepackage{graphicx}
\usepackage{comment}
\usepackage{amsmath,amssymb} % define this before the line numbering.
\usepackage{color}

\usepackage{subfigure}
\usepackage{enumitem}
\usepackage{makecell}
\usepackage{multirow}
\usepackage{seqsplit}

\newcommand{\modelname}[0]{F-PAD}
\DeclareMathOperator*{\argmax}{arg\,max}
\newcommand{\ssc}{\sigma} % similarity score

\begin{document}
\pagestyle{headings}
\mainmatter
\def\ECCVSubNumber{****}  % Insert your submission number here

\title{Revisiting Few-shot Activity Detection with Class Similarity Control}

%\begin{comment}
%\titlerunning{ECCV-20 submission ID \ECCVSubNumber} 
%\authorrunning{ECCV-20 submission ID \ECCVSubNumber} 
%\author{Anonymous ECCV submission}
%\institute{Paper ID \ECCVSubNumber}

\titlerunning{Revisiting Few-shot Activity Detection with Class Similarity Control}
\author{Huijuan Xu$^1$ \hspace{1.5mm} 
Ximeng Sun$^2$  \hspace{1.5mm}
Eric Tzeng$^1$ \hspace{1.5mm} 
Abir Das$^3$ \hspace{1.5mm}
Kate Saenko$^2$  \hspace{1.5mm}
Trevor Darrell$^1$%\\
}
\authorrunning{Huijuan Xu et al.}
\institute{$^1$University of California, Berkeley  %\qquad
$^2$Boston University
$^3$IIT Kharagpur \\
\email{%\tt\small 
$^1$\{huijuan, etzeng, trevor\}@eecs.berkeley.edu, \\
$^2$\{sunxm, saenko\}@bu.edu, 
$^3$abir@cse.iitkgp.ac.in}
}%\\

%\end{comment}
%******************

\begin{comment}
\titlerunning{Abbreviated paper title}
\author{First Author\inst{1}\orcidID{0000-1111-2222-3333} \and
Second Author\inst{2,3}\orcidID{1111-2222-3333-4444} \and
Third Author\inst{3}\orcidID{2222--3333-4444-5555}}
%
\authorrunning{F. Author et al.}
\institute{Princeton University, Princeton NJ 08544, USA \and
Springer Heidelberg, Tiergartenstr. 17, 69121 Heidelberg, Germany
\email{lncs@springer.com}\\
\url{http://www.springer.com/gp/computer-science/lncs} \and
ABC Institute, Rupert-Karls-University Heidelberg, Heidelberg, Germany\\
\email{\{abc,lncs\}@uni-heidelberg.de}}
\end{comment}
%******************
\maketitle

\begin{abstract}
Many interesting events in the real world are rare making preannotated machine learning ready videos a rarity in consequence.
Thus, temporal activity detection models that are able to learn from a few examples are desirable. In this paper, we present a conceptually simple and general yet novel framework for few-shot temporal activity detection based on proposal regression which detects the start and end time of the activities in untrimmed videos. Our model is end-to-end trainable, takes into account the frame rate differences between few-shot activities and untrimmed test videos, and can benefit from additional few-shot examples.
We experiment on three large scale benchmarks for temporal activity detection (ActivityNet1.2, ActivityNet1.3 and THUMOS14 datasets)~in a few-shot setting.
We also study the effect on performance of different amount of overlap with activities used to pretrain the video classification backbone and propose corrective measures for future works in this domain.
Our code will be made available.
\keywords{Few-shot Temporal Activity Detection}
\end{abstract}

%\vspace{10pt}
\section{Introduction}
The rising popularity of video recording devices and social media has led to an explosion of available video data.
Automatic analysis of such data is challenging because most of the videos are in untrimmed form and contain very few interesting events. 
Temporal activity detection tries to automatically detect the start and end time of interesting events and enables many real-world applications, \textit{e.g.}, unusual event detection in surveillance video.
The fact that these events are rare requires models to detect them based on a few training examples.

What should an efficient few-shot activity detection model look like?
Unlike the large data scenario detecting activities over the whole activity label space, the few-shot setting in real applications should be able to accept new out-of-distribution examples and generate reasonable predictions.
Thus, the predicted activity class in few-shot activity detection setting needs to be based on similarities between few-shot activity representations and interesting candidate segments in the video.
To realize this, few-shot activity detection is performed as example-based activity detection where one stream encodes the few-shot activity videos along with a main stream that encodes the untrimmed video and extracts candidate proposals likely to contain activities.
A final label is put to the candidate proposals depending on the similarity between the query few-shot videos and the proposed candidates.
Yang \textit{et. al.}~\cite{yang2018one} address few-shot activity detection via example-based action detection based on the Matching Network~\cite{vinyals2016matching}. 
It utilizes correlations between the sliding window feature encoding and few-shot example video features, to localize actions of previously unseen classes.
However, due to the use of ``sliding windows'' this approach suffers from high computational cost and inflexible activity boundaries.

\begin{figure}[t]
    \centering
    \includegraphics[width=0.95\linewidth]{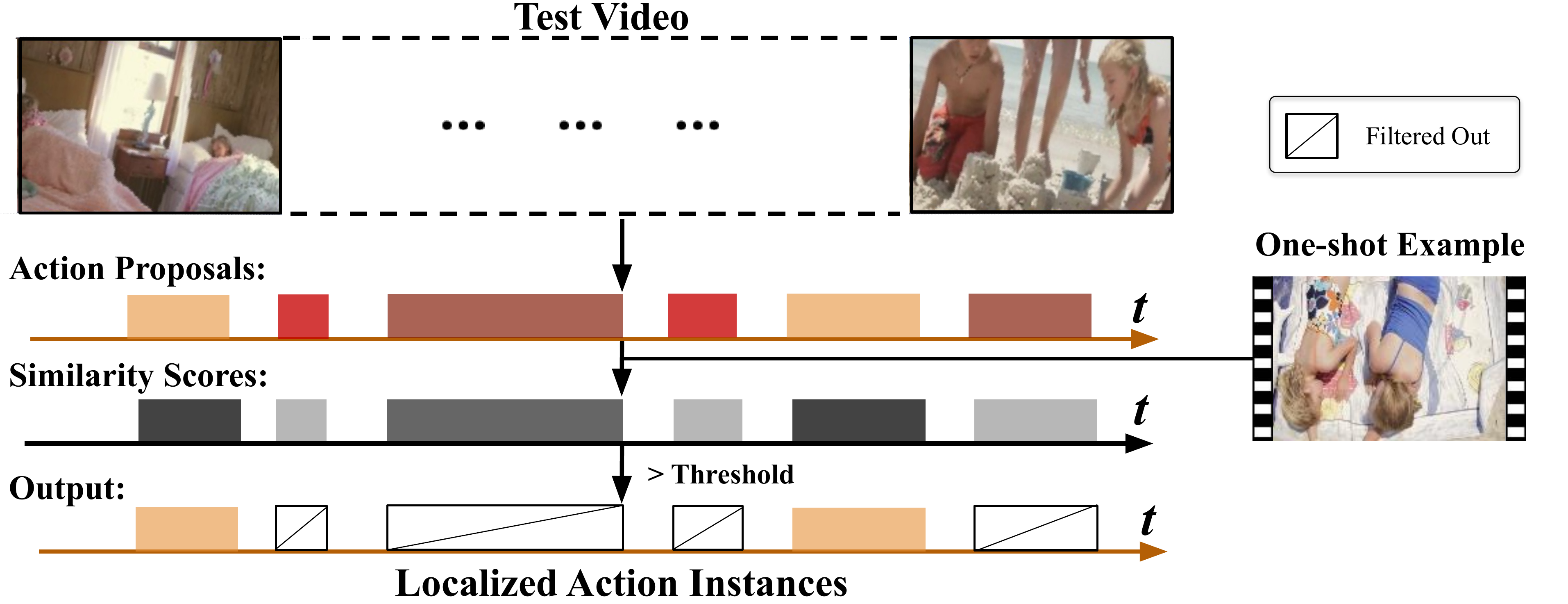}
    %\vspace{5pt}
    \caption{Overview of our proposal regression based few-shot activity detection model in untrimmed video.
    Our model proposes foreground segments (action proposals) and assigns labels to them based on their similarity to the one/few-shot training examples of activities.
    Additionally, we generate proposal features adaptive to frame rate differences between proposal stream and untrimmed video stream.
    }
    \label{fig:problem_overview}
    \vspace{-10pt}
\end{figure}

To generate more accurate and efficient proposals for few-shot detection, we propose an approach based on direct coordinate regression for temporal localization, called Few-shot Proposal-based Activity Detection model (\textit{\modelname}), shown in Fig.~\ref{fig:problem_overview}.
We encode both few-shot examples and the untrimmed video using 3D convolutional (C3D) features~\cite{tran2015learning} followed by a temporal proposal network extracting action proposals from the untrimmed video.
We produce similarity scores by comparing each few-shot example with the proposed candidates in the encoded feature space and assign each proposal an activity label same as that of the maximally similar few-shot video.
To the best of our knowledge, the proposed F-PAD model is the first proposal based few-shot activity detection approach.

We also notice that there can be discrepancies in the encoded feature space between the action proposals and the few-shot videos due to different frame rates between them.
As far as the proposals are concerned, they are extracted by segment of interest pooling~\cite{xu2017r} from untrimmed videos with fixed frame-rate.
However, the few-shot activity videos can be of varied duration and thus fixed number of frames are sampled from them resulting in a possibly different frame rate compared to the action proposals.
To minimize the feature discrepancy arising from these frame rate differences, we design a distributional adaptation loss during training which we show, leads to improved performance.
Our model is end-to-end trainable, and demonstrates the advantages of proposal localization and feature learning in an end-to-end fashion.
Our experiments also reveal that increasing the number of training videos per class have larger relative improvement over the existing state-of-the-art~\cite{yang2018one}.

Traditionally, an action/object detection model is warmstarted from a pre-trained classification backbone.
However, in few-shot action/object detection task, this practice can result in potential overlap between the novel classes for detection and the activity classes seen by the pre-trained classification backbone.
Still, few-shot object detection~\cite{dong2018few,kang2019few} uses pre-trained imagenet classification backbone riding on the justification that the labels used in classification and detection are different.
As detection requires bounding box annotations in addition to the class labels, the argument is that detection still remains few shot even if the classification backbone is trained on a large-scale classification dataset.
In this paper, we investigate the effect of the potential class overlap between training classes in the pre-trained activity classification model and the few-shot activity detection model.
The analysis is done by creating different random splits of train (base classes) and test (novel classes) data where the overlap of novel classes with large-scale pretraining dataset is a control variable.
Based on the experiments, we conclude that traditional evaluation of few-shot activity detection approaches are grossly overestimated.
It does not take into consideration possible leakage of information from large-scale pre-training where training data for the pre-trained (classification) model and the test data of the target (detection) model may have significant overlap.
As a remedy, we propose to evaluate the performance more carefully with absolutely disjoint train and test classes in pre-training and target model respectively.

%\vspace{5pt}
To summarize, our contributions are:
%\vspace{-1.5pt}
\begin{itemize} [leftmargin=*]
    \itemsep0em
    \item We propose a few-shot activity detection model (\textit{\modelname}) with an end-to-end trainable proposal network to detect potential events efficiently and concurrently;
    \item We introduce a novel distributional adaptation loss to minimize the feature discrepancy from different frame rates;
    \item We evaluate our model on different base and novel splits of three activity detection datasets with different class overlaps and demonstrate the effect of such overlap in few-shot activity detection performance. We hope our simple and effective approach will improve our understanding behind the true reason of improvement for few-shot activity detection approaches, and help ease future research in evaluating few-shot activity detection.
\end{itemize}

%\vspace{10pt}
\section{Related Work}
\label{Related}

%%%%%%%%%%%%%%%%%%%%%%%%%%%%%%%%%%%%%%%%%%%%%%%%%%%%%%%%%%%%%%%%%%%%%%%%
%\vspace{10pt}
\subsection{Activity Detection}
We focus on temporal activity detection~\cite{escorcia2016daps,ma2016learning,montes2016temporal,shou2016temporal,singh2016multi,yeung2016end} in this paper which predicts the start and end times of the activities and classifies them in long untrimmed videos.
Existing temporal activity detection approaches are dominated by models that use sliding windows to generate segments and subsequently classify them with activity classifiers trained on multiple features~\cite{karaman2014fast,oneata2014lear,shou2016temporal,wang2014action}.
The use of exhaustive sliding windows is computationally inefficient and constrains the boundary of the detected activities to some extent.
Later, some approaches have bypassed the need for exhaustive sliding window search and proposed to detect activities with arbitrary lengths~\cite{escorcia2016daps,ma2016learning,montes2016temporal,singh2016multi,yeung2016end}.
This is achieved by modeling the temporal evolution of activities using RNNs or LSTM networks and predicting an activity label at each time step.
Recently, CDC~\cite{shou2017cdc} and SSN~\cite{zhao2017temporal} propose bottom-up activity detection by first predicting labels at the frame/snippet-level and then fusing them.
The R-C3D activity detection pipeline~\cite{xu2017r} encodes the frames with fully-convolutional 3D filters, finds activity proposals, then classifies and refines them based on pooled features.
We use a similar proposal network as is used in this model and adapt it to for our few-shot detection scenario.

%\vspace{10pt}
\subsection{Few-shot Learning}
The goal of few-shot learning is to generalize to new examples despite few labeled training data. 
The conventional setting tests on novel classes only, while a generalized setting tests on both seen and unseen classes.
Metric learning methods based on measuring similarity to few-shot example inputs are widely employed by many few-shot learning algorithms to good effect~\cite{koch2015siamese,shyam2017attentive,snell2017prototypical,vinyals2016matching}.
Similarity values are typically defined on selected attributes or word vectors~\cite{mishra2018generative,zellers2017zero}.
Another line of few-shot learning algorithms augments data by synthesizing examples for the unseen labels~\cite{long2017zero,zhang2018visual}.
Other approaches train deep learning models to learn to update the parameters for few-shot example inputs~\cite{andrychowicz2016learning,ravi2016optimization,bertinetto2016learning}.

Recently, most few-shot models tend to utilize the meta-learning framework for training on a set of datasets with the goal of learning transferable knowledge among datasets~\cite{santoro2016one,mishra2017meta}. MAML~\cite{finn2017model} is one typical example which combines the meta-learner and the learner into one, and directly computes the gradient with respect to the meta-learning objective.
\cite{mettes2017spatial} studies the zero-shot spatial-temporal action localization using spatial-aware object embeddings.
\cite{yang2018one} is the first work proposing the task of temporal activity detection based on few-shot input examples. It performs activity retrieval for the few-shot example(s) through a sliding window approach and then combines bottom-up retrieval results to obtain the final temporal activity detection.
This work follows the conventional setting of testing on novel examples only, and uses MAML training strategy.
In our paper, we work along this direction of example based few-shot temporal activity detection, and propose a proposal based few-shot activity detection model for accurate and efficient temporal localization.
Additionally, we also show the effect of controlling the novel class similarity with the pre-training classes on few-shot detection performance.

\begin{figure*}[t]
    \centering
    \includegraphics[width=\linewidth]{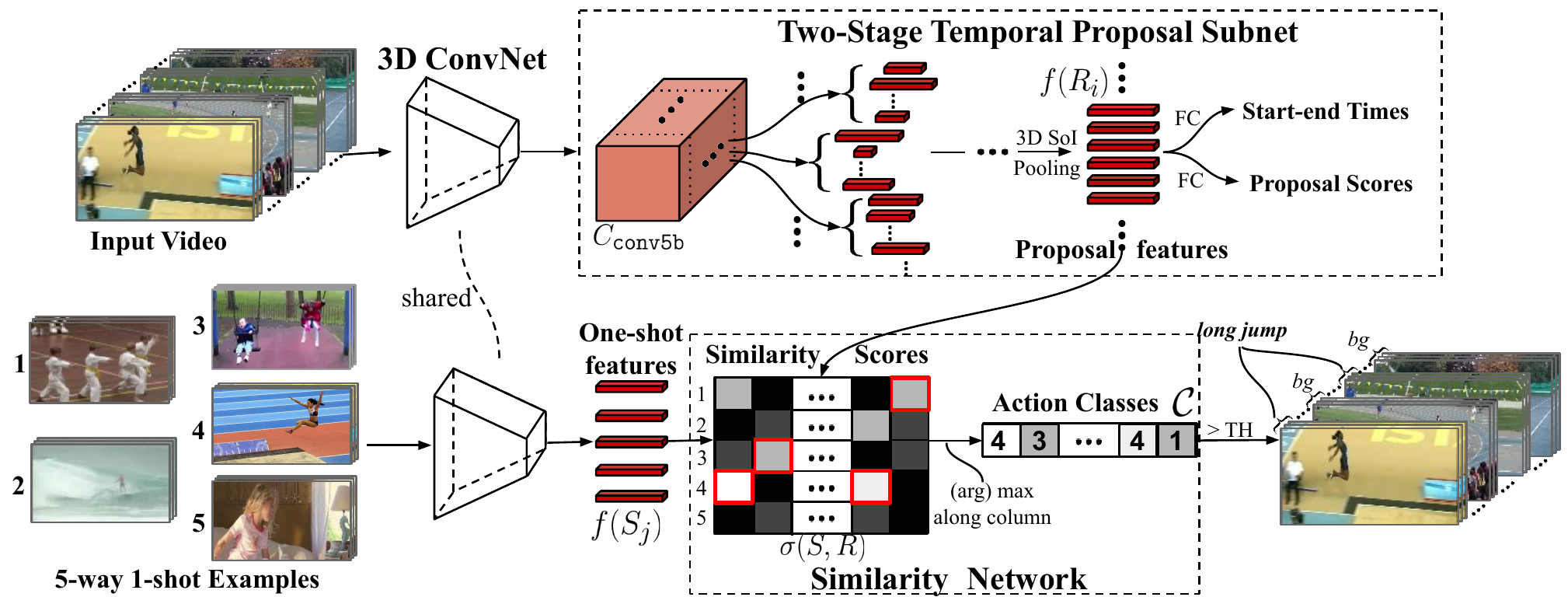}
    %\vspace{10pt}
    \caption{Detail diagram of our proposed F-PAD network for exampled based few-shot activity detection. Given an untrimmed test video, and one- or few-shot training examples of an activity class, the network detects instances of the given activities in the test video and localizes their start and end times. The model has three components: (1) a 3D ConvNet feature extractor applied to both the untrimmed input video and the few-shot trimmed videos, (2) a temporal proposal network, and (3) a similarity network that computes similarity scores between proposals and few-shot videos and assigns labels.}
    \label{fig:architecture}
    %\vspace{10pt}
\end{figure*}

%\vspace{10pt}
\section{Example based Temporal Activity Detection}
\label{sec:approach}

In this paper, we tackle the task of few-shot temporal activity detection.
The goal is to locate all instances of activities (temporal segments) in an untrimmed test video, given only a few typical examples from a new set of activity classes available for training.
Ideally, a few-shot temporal activity detection model should be equipped with a proposal regression module which can benefit from end-to-end training. It also needs to conduct few-shot example matching with regressed temporal proposals and address the frame rate difference between few-shot activities and untrimmed videos.

Based on these thoughts, we propose a two-stream end-to-end model with proposal regression,  called \textit{Few-shot Proposal based Activity Detection model (F-PAD)} to tackle this task.
Our key idea is to assign each detected proposal with one of the few-shot class labels  based on the maximum similarity score between features from each proposal and few-shot example videos.

A detailed diagram of our F-PAD network is in Figure~\ref{fig:architecture}.
The model consists of three components: a 3D ConvNet feature extractor~\cite{tran2015learning}, a temporal proposal network, and a similarity network.
Specifically, the 3D ConvNet feature extractor encodes both the untrimmed input video and the few-shot trimmed example videos; the proposal subnet predicts temporal segments of variable lengths that contain potential activities; the similarity network classifies these proposals into one of the few-shot activity categories.
The feature discrepancy between proposals and few-shot trimmed videos caused by different frame rates is compensated during training by adding a novel distributional loss.
Details are described in next few sections.

%\vspace{10pt}
\subsection{Two-Stream Video Feature Encoding}
\label{sec:feature}
Addressing few-shot temporal activity detection as example-based detection, requires us to design two streams to accept two types of input videos, namely, the untrimmed video and the trimmed few-shot example videos.
We employ a 3D ConvNet~\cite{tran2015learning} to extract rich spatio-temporal feature hierarchies for our F-PAD model. 
For the few-shot branch in the bottom of Figure~\ref{fig:architecture}, we take five trimmed videos belonging to five classes, with one video from each class in the one-shot setting.
In general, for a $N$-way, $k$-shot scenario videos from $N$ classes are taken where $k$ is the number of videos in each class.
We uniformly sample $L=16$ frames from each few-shot video, and then get a fixed-dimensional feature vector $f(S_j)$  for each video $j$ via the 3D ConvNet (see ``One-shot features'' in Figure~\ref{fig:architecture}).

For the untrimmed input video in the upper branch, we use fully convolutional layers in C3D to encode a long sequence of frames (sampled at a fixed frame rate) resulting in the feature map~$C_{conv5b}$.
From the untrimmed video feature encoding~$C_{conv5b}$, our proposal subnet (Sec.~\ref{sec:proposal}) proposes candidate segment proposals and extracts proposal features $f(R_i)$ for each proposal $i$.
The two video encoding branches share the same 3D ConvNet weights.
We further minimize the feature encoding discrepancy arising from the apparent frame rate differences between inputs to these two branches.
Finally, similarities are measured between proposal features $f(R_i)$ and few-shot video feature encoding $f(S_j)$ for few-shot classification (see Sec.~\ref{sec:similarityNetwork}).

%\vspace{10pt}
\subsection{Temporal Proposal Subnet}
\label{sec:proposal}
We inherit the temporal proposal module from the R-C3D temporal activity detection model~\cite{xu2017r}, and additionally, we adapt the second classification module in R-C3D to be class agnostic and form a two-stage temporal proposal subnet, based on the previous experience that refined temporal proposals from second stage are generally better than the initial proposals from first stage.
The input to the temporal proposal network is the untrimmed video feature encoding $C_{conv5b}$.
After applying a series of 3D convolutional filters and 3D max-pooling layers to $C_{conv5b}$, it predicts the temporal proposals with respect to a set of anchor segments.
Each temporal proposal is also paired with a binary label (score) that indicates whether the proposal is a foreground activity or background.
High quality temporal proposals are selected to extract temporal proposal features~$f(R_i)$ from the untrimmed video features $C_{conv5b}$ by 3D Segment of Interest (SoI) pooling~\cite{xu2017r} (shared with Proposal Stage I).
Two separate fully-connected heads (a classification and a regression layer) classify the proposal~$i$ into foreground/background and predict refined start and end time respectively.
For each proposal $i$, the features $f\left( R_i \right)$ extracted from the aforementioned fully-connected layers after 3D SoI pooling layer are subsequently used in the Similarity Network to compute similarities with the few-shot video features $f\left( S_j \right)$ (See Sec.~\ref{sec:similarityNetwork}).

%\vspace{10pt}
\subsection{Similarity Network for Few-shot Classification}
\label{sec:similarityNetwork}
Unlike the temporal activity detection which directly classifies over the whole activity label space using the proposal features $f\left( R_i \right)$, in few-shot setting, we can't anticipate the few-shot labels when we pre-define the activity label space in real application.
Thus in example-based few-shot setting, the activity classification is changed to measure the feature similarities between features of the proposal $i$ and the few-shot input video $j$ (i.e $f\left( R_i \right)$ and $f\left(S_j\right)$).
A Similarity Score matrix $\ssc\left(S,R\right)$ is calculated between  such $f\left( R_i \right)$s and $f\left(S_j\right)$s.
The $ij^{th}$ element $\ssc\left(S_j,R_i\right)$ is the cosine similarity between $f\left(S_j\right)$ and $f\left(R_i\right)$:

\begin{equation}
\ssc(S_j,R_i)=\frac{ \big \langle f \big(S_j \big),f \big(R_i \big) \big \rangle}{\|f \big (S_j \big)\|\|f \big(R_i \big)\|}.
\end{equation}

Each proposal $i$ is assigned a few-shot class label $\mathcal{C}_i$, based on the Similarity Score matrix $\ssc\left(S,R\right)$ between the proposal features and the few-shot video features.
In the one-shot setting, $\argmax$ is directly applied on each column of the Similarity Score matrix $\ssc(S,R)$: 
\begin{equation}
    \mathcal{C}_i=\argmax_j\ssc \big(S_j,R_i \big).
\end{equation}
In multi-shot setting, we first average the similarity scores of multiple examples belonging to the same class, and then assign $\mathcal{C}_i$ by applying $\argmax$ among the few-shot class labels.

%\vspace{10pt}
\subsection{Frame Rate Adaptation}
\label{sec:frame_rate}
As mentioned in section~\ref{sec:feature},  there is inherent frame rate difference between the few-shot videos and the untrimmed videos in our two stream video feature encoding: 
the model samples fixed number frames uniformly from few-shot videos with variable lengths, while it samples from untrimmed videos using fixed frame rate.
This problem potentially affects the video feature encoding for similarity calculation.
To compensate for the feature discrepancy brought by different frame rates, we borrow a domain adaptation loss $L_{adaptation}$ to supervise the feature learning for temporal proposals and few-shot trimmed videos:
\begin{equation}
    L_{adaptation} = \left \| \frac{1}{N} \sum_i^N f(R_i) - \frac{1}{M} \sum_j^M f(S_j) \right \|_2,
\end{equation}
where the temporal proposal features~$f(R_i)$ are averaged over all the $N$ proposals and the few-shot video features~$f(S_j)$ are averaged over all the $M$ few-shot input videos, and L2 penalty is utilized to minimize the distributional discrepancy between $f(R_i)$ and $f(S_j)$ and learn frame rate adaptive features.

%\vspace{10pt}
\subsection{Optimization}
\label{sec:optimization}
Beside the adaptation loss ${L}_{adaptation}$ introduced in Sec.~\ref{sec:frame_rate}, the training loss~${L}_{total}$ for F-PAD contains the losses from the two proposal stages~(${L}_{p1}$ and ${L}_{p2}$) and the few-shot Similarity Network~(${L}_{fewShot\_cls}$), resulting in the following total loss:
\small
\begin{equation}
    {L}_{total} = {L}_{p1} + {L}_{p2} + {L}_{fewShot\_cls} + \lambda {L}_{adaptation}.
%\vspace{-2mm}
%\label{eq:overal_loss}
\end{equation}
%\normalsize

Each of the proposal losses ${L}_{p1}$ and ${L}_{p2}$ consists of a smooth L1 regression loss~\cite{girshick2015fast} for proposal regression and a binary classification loss for binary proposal classification.
The few-shot classification loss~${L}_{fewShot\_cls}$ in the few-shot Similarity Network is a multi-class cross-entropy loss.
$\lambda$ indicates the proportion of the distributional adaption loss ${L}_{adaptation}$ in the total loss.
When constructing the training batch for the two proposal stages as well as the Similarity Network, we sample batches balanced with a fixed positive to negative ratio, and the positive/negative examples are determined by temporal Intersection-over-Union (tIoU) thresholds.

%%%%%%%%%%%%%%%%%%%%%%%%%%%%%%%%%%%%%%%%%%%%%%%%%%%%%%

%\vspace{10pt}
\subsection{Few-shot Activity Detection}
\label{sec:prediction}

Few-shot temporal activity detection in F-PAD model consists of two steps.
First, the temporal proposal network generates candidate proposals by predicting the start-end time offsets as well as a proposal score for each.
After NMS with IoU threshold 0.7, the selected proposals from the first stage are fed to the second stage to perform binary proposal classification again, and the boundaries of the selected proposals are further refined by a regression layer.
Second, proposals from the second stage are used to calculate similarity scores with features from few-shot trimmed videos, and each proposal is assigned the few-shot class label with the maximum similarity score.
In multi-shot scenario, the scores are first averaged within each class, and then max is taken among few-shot classes in that episode.
Finally, each proposal is paired with a proposal score and a few-shot class with maximum similarity score, and the detected activity is determined by the use of a threshold on these scores.

%\vspace{10pt}
\section{Experiments}

\subsection{Evaluation Setup}
We use the same few-shot experiment setup as in the previous work~\cite{yang2018one}, and split the classes in each dataset into two disjoint subsets: base classes for training and novel classes for testing.
The specifics about the splits for each dataset are detailed in the corresponding sections for these datasets.
We also use the same meta-learning setup as in \cite{yang2018one}.
During the training stage, in each training iteration, we randomly sample five classes from the subset of training classes, and then for each class we randomly sample one example video (in the one-shot setting) to constitute an input training batch to the few-shot branch.
In the five-shot setting, we randomly sample five example videos for each class, and average the scores of five example videos in same class to get the final similarity value for that class.
Notably, the randomly sampled five classes for the few-shot branch should have at least one class overlapped with the ground truth activity classes in the randomly sampled untrimmed input video.
The five sampled activity classes are assigned labels from 0-4 randomly.

The testing data preparation strategy is also same as that of the aforementioned training stage and the test data comes from the subset of testing classes.
Results are shown in terms of mean Average Precision - mAP@$\alpha$ where $\alpha$ denotes Intersection over Union (IoU) threshold, and average mAP at 10 evenly distributed IoU thresholds between 0.5 and 0.95, as is the common practice in the literature.
However in the testing stage, mAP@0.5 and average mAP are calculated in each iteration for one untrimmed test video, and the final results are reported by averaging over 1000 iterations.

%%%%%%%%%%%%%%%%%%%%%%%%%%%%%%%%%%%%%%%%%%%%%%%%%%%%%%%%%%%%%%%%%%%%%%%%%%%%%%%%
\subsection{Experimental Setup}
\label{exp:different_setup}
In this paper, we evaluate \modelname~on three large-scale activity detection datasets - ActivityNet1.2 and ActivityNet1.3~\cite{caba2015activitynet}, and THUMOS'14~\cite{THUMOS14}.
Next, we introduce the experimental setup on these three datasets.

%\vspace{3mm}
%\noindent
\textbf{ActivityNet1.2}
The ActivityNet~\cite{caba2015activitynet} dataset consists of untrimmed videos and has three versions.
The ActivityNet1.2 dataset contains around 10k videos and 100 activity classes which are a subset of the 200 activity classes in ActivityNet1.3.
Most videos contain activity instances of a single class covering a great deal of the video.
In the ActivityNet1.2 dataset, the trimmed videos in the few-shot branch come from the ground truth annotations inside each untrimmed video.
The length of the input video buffer is set to 768 at 6 fps.
As a speed-efficiency trade-off, we freeze the first two convolutional layers in our C3D model during training and the learning rate is fixed at $10^{-4}$.
The proposal score threshold is set as 0.3.
We initialize the 3D ConvNet with C3D weights trained on Sports-1M released by the authors in~\cite{tran2015learning} for all the three datasets.
This classification backbone has similar classification accuracy with the video classification backbone used in the previous work~\cite{yang2018one} which is roughly 80\% on UCF-101 dataset.
We pre-train the temporal proposal network using the proposal annotations from ActivityNet1.2 train data.
Regarding the base and novel data splits, the 100 activity classes are randomly split into 80 classes (ActivityNet1.2-train-80) for training and 20 classes (ActivityNet1.2-test-20) for testing, and we do this random split three times resulting in ``randomSplit1", ``randomSplit2", ``randomSplit3".
These splits are used to test and evaluate the performance when there can be possible overlap between pretraining classification classes and novel detection classes.
Next, to decouple any overlap, we manually check the 100 activity classes and pick the classes those do not exist in the Sports-1M dataset.
We found 63 such activity classes.
Now, we randomly sample 20 classes (ActivityNet1.2-test-20) for testing from these 63 classes, and the rest of 80 classes (ActivityNet1.2-train-80) are used for training.
We perform this controlled random split three times resulting in ``controlled\_randomSplit1", ``controlled\_randomSplit2", ``controlled\_randomSplit3".
We report results on these six different base and novel splits.

%\vspace{3mm}
%\noindent
\textbf{ActivityNet1.3}
The 200 activity classes in ActivityNet1.3 dataset are randomly split into 180 classes (ActivityNet1.3-train-180) for training and 20 classes (\seqsplit{ActivityNet1.3-test-20}) for testing. 
Other settings (the buffer length, pretraining, learning rate, etc.) are kept same as those in the ActivityNet1.2 dataset. 
The proposal score threshold is set as 0.5.
Regarding the base and novel data splits, we follow the same strategy as in ActivityNet1.2.
``randomSplit1", ``randomSplit2", ``randomSplit3" denotes the random split where there is no control on the possible overlap between pretraining classification classes and novel detection classes.
Similarly, the three controlled random splits (with no overlap between pretraining classes and novel classes) are denoted as ``controlled\_randomSplit1", ``controlled\_randomSplit2", ``controlled\_randomSplit3".
Note that we keep the classes in ActivityNet1.2-test-20 and ActivityNet1.3-test-20 the same.

%%%%%%%%%%%%%%%%%%%%%%%%%%%%%%%%%%%%%%%%%%%%%%%%%%%%%%%%%
\begin{figure*}[t]
\centering
    \centering
    \includegraphics[width=0.98\linewidth]{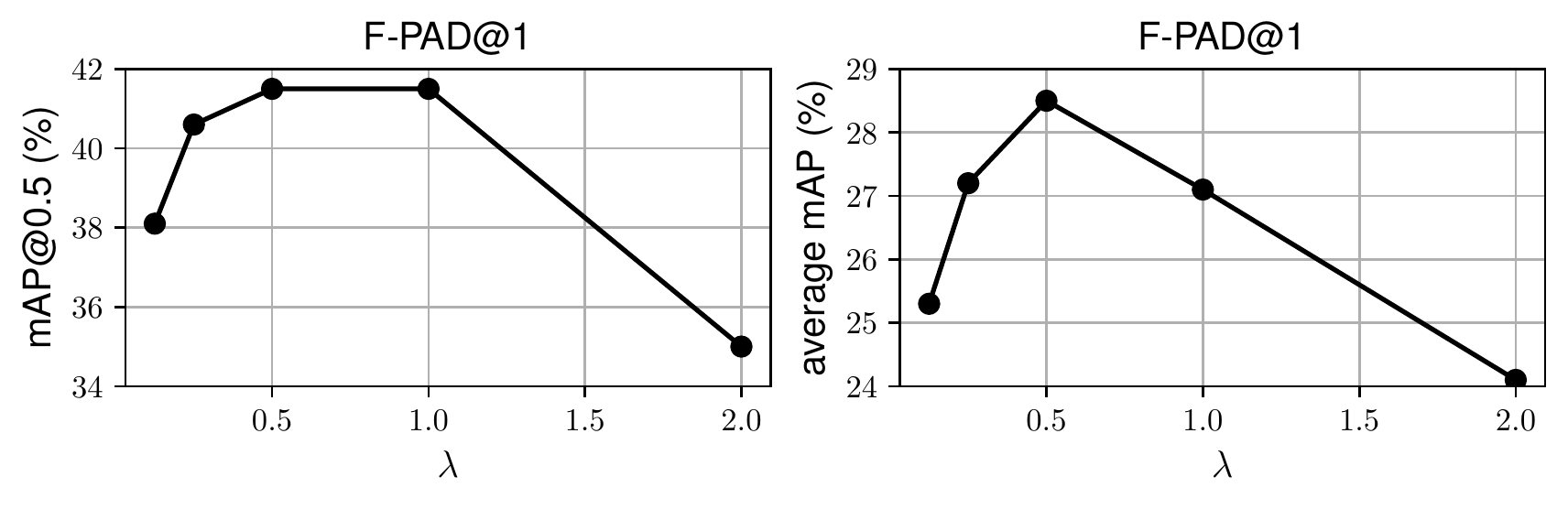} 
    \caption{Ablations about different adaptation loss weights $\lambda$ for minimizing frame rate differences on ActivityNet1.2. mAP at tIoU threshold $\alpha=0.5$ and average mAP of $\alpha \in \{0.5,0.95\}$ are reported (in percentage). @1 means ``one-shot".  
    }
    \label{fig:res_act100_frameRate}
    %\vspace{10pt}
\end{figure*}
%%%%%%%%%%%%%%%%%%%%%%%%%%%%%%%%%%%%%%%%%%%%%%%%%%%%%%%%%
\begin{figure}[t]
    \centering
    \includegraphics[width=\linewidth]{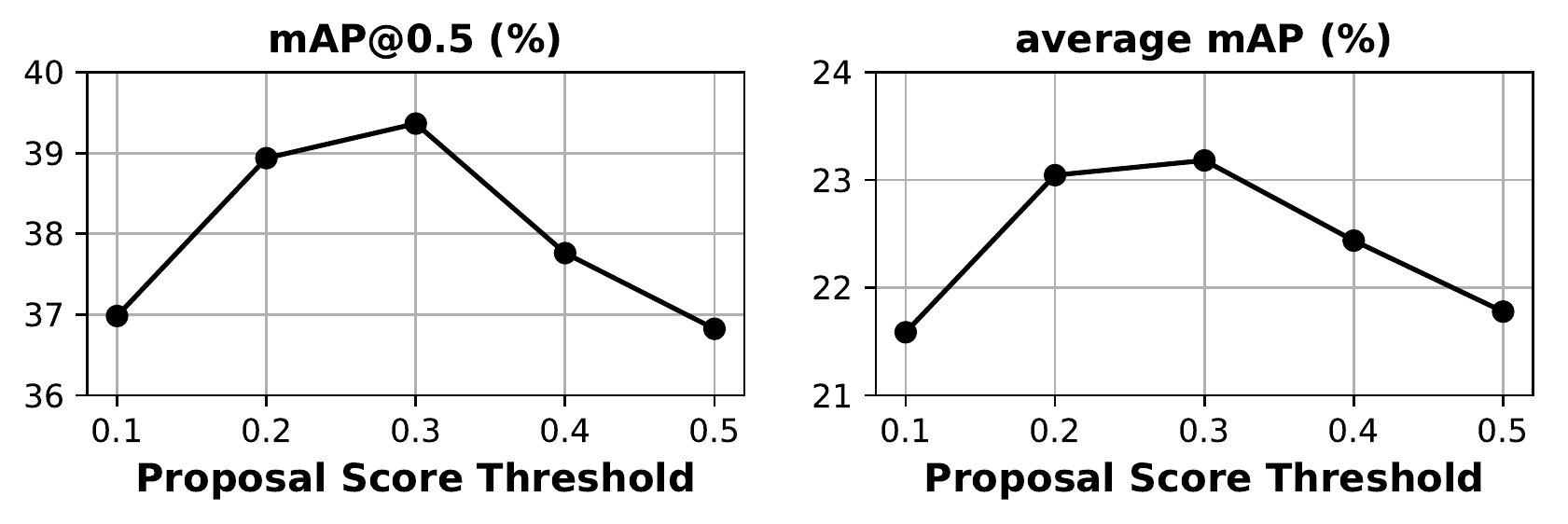}
    \caption{The change of mAP@0.5 and average mAP (in percentage) with the proposal score threshold for the F-PAD model in one-shot setting on ActivityNet1.2 dataset.}
    %\vspace{5pt}
    \label{fig:mAP_vs_thres}
   %\vspace{5pt}
\end{figure}
%%%%%%%%%%%%%%%%%%%%%%%%%%%%%%%%%%%%%%%%%%%%%%%%%%%%%%%%%

\textbf{THUMOS'14}
The THUMOS'14 activity detection dataset contains videos of 20 different sport activities.
The training set contains 2765 trimmed videos from UCF101 dataset~\cite{soomro2012ucf101} while the validation and test sets contain 200 and 213 untrimmed videos respectively.
The 20 sport activities in THUMOS'14 are a subset of the 101 classes in UCF101.
In the THUMOS'14 dataset, the trimmed videos in the few-shot branch of our model come from UCF101.
The input video buffer is set as 512 frames at 25 fps.
We allow all the layers to be trained with a fixed learning rate of $10^{-4}$.
The proposal score threshold is set as 0.05 and the similarity score threshold as 0.02. 
Regarding the base and novel data splits, we first follow previous work~\cite{yang2018one} to split the 20 classes into 6 classes (Thumos-val-6) for training and 14 classes (Thumos-test-14) for testing, and do the random split three times without class similarity control resulting in ``randomSplit1", ``randomSplit2", ``randomSplit3".
Then we checked the class overlap between 20 classes in THUMOS'14 and the 487 classes in Sports-1M.
Out of the 20 classes, 9 classes do not exist in Sports-1M.
Thus we could create only one controlled split with 11 classes (Thumos-val-11) for training and 9 classes (Thumos-test-9) not existing in Sports-1M for testing, resulting in ``controlled\_randomSplit1".

\subsection{Model Ablation on ActivityNet1.2}
\begin{figure*}[t]
\centering
    \centering
    \includegraphics[width=0.98\linewidth]{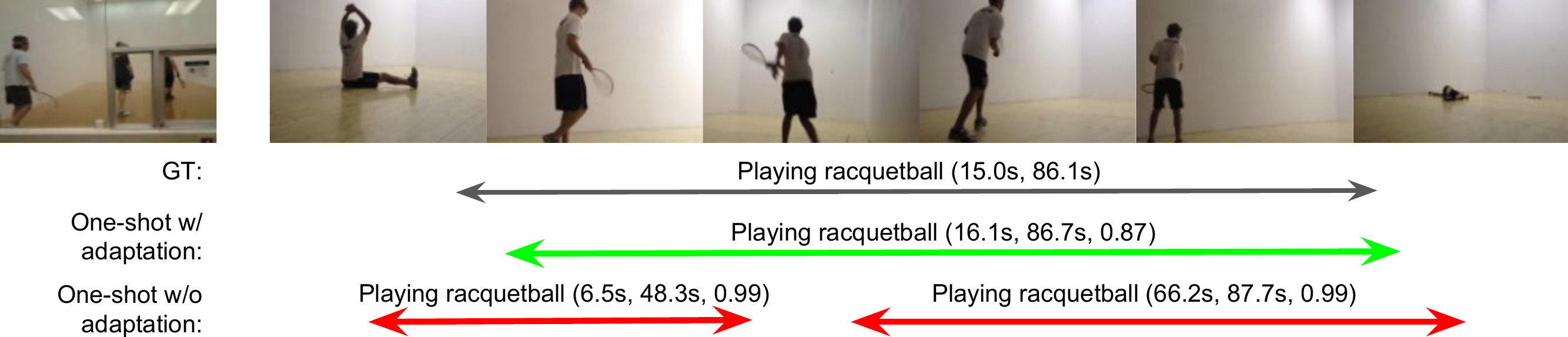} \\ ~ \\
    %\vspace{10pt}
    \includegraphics[width=0.98\linewidth]{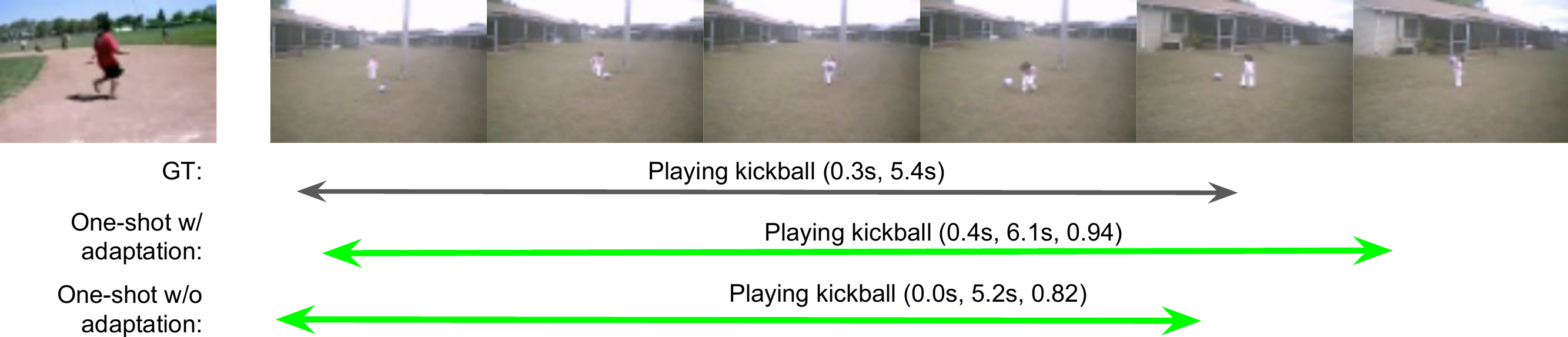}
    %\vspace{10pt}
    \caption{Qualitative visualization of one-shot temporal activity detection results from our~\modelname~model with and without adaptation loss on activityNet1.2 dataset. 
    Two examples are shown.
    For each example we first show the one-shot input video and the untrimmed video, then ground truth (GT) annotation, prediction with adaptation and prediction without adaptation are shown in each row in sequence.
    GT annotation are marked with black arrows. Correct predictions (predicted clips having tIoU more than 0.5 with ground truth) are marked in green, and incorrect predictions are marked in red. The start and end times are shown in seconds as well as the similarity score. Best viewed in color.}
    \label{fig:act100_examples_adaptation}
    %\vspace{10pt}
\end{figure*}

%\vspace{10pt}
\paragraph{Ablation Study for the Adaptation Loss of Frame Rates:}
In this paper, we consider the frame rate differences in our two input branches and incorporate the distributional level adaptation loss $L_{adaptation}$ to minimize the feature discrepancy brought in by different frame rates.
We experiment with different loss weight $\lambda$ for the adaptation loss $L_{adaptation}$, and mAP@0.5 and average mAP results are shown in Figure~\ref{fig:res_act100_frameRate}.
$\lambda=0.5$ gives the best adaptation performance in one shot setting.
We show some example predictions from one-shot models with adaptation and without adaptation in Figure~\ref{fig:act100_examples_adaptation}.
It can be seen that our model with adaptation performs better for long duration activities while the model without adaptation is worse.

%%%%%%%%%%%%%%%%%%%%%%%%%%%%%%%%%%%%%%%%%%%%%%%%%%%%%%%%%
%%%%%%%%%%%%%%%%%%%%%%%%%%%%%%%%%%%%%%%%%%%%%%%%%%%%%%%%%

%\vspace{7pt}
\paragraph{Ablation Study for Proposal and Similarity Scores:}
Recall that our final few-shot detection results are determined by thresholding over the proposal score.
To show the effect of the proposal score threshold, we plot mAP@0.5 and average mAP with respect to the proposal score threshold for our one-shot setting on the ActivityNet1.2 dataset in Figure~\ref{fig:mAP_vs_thres}.
From the figures, we can see that there is a trade-off between the quantity and the quality of proposals, and in this setting the optimal proposal score threshold is 0.3.
The similarity score mainly takes the role of assigning activity label to proposals by taking maximum in each episode and the corresponding similarity score threshold doesn't have too much effect.

\subsection{One-shot Results on Different Base and Novel Data Splits}
\label{exp:different_split}
\begin{figure*}[t]
\centering
    \centering
    \includegraphics[width=0.98\linewidth]{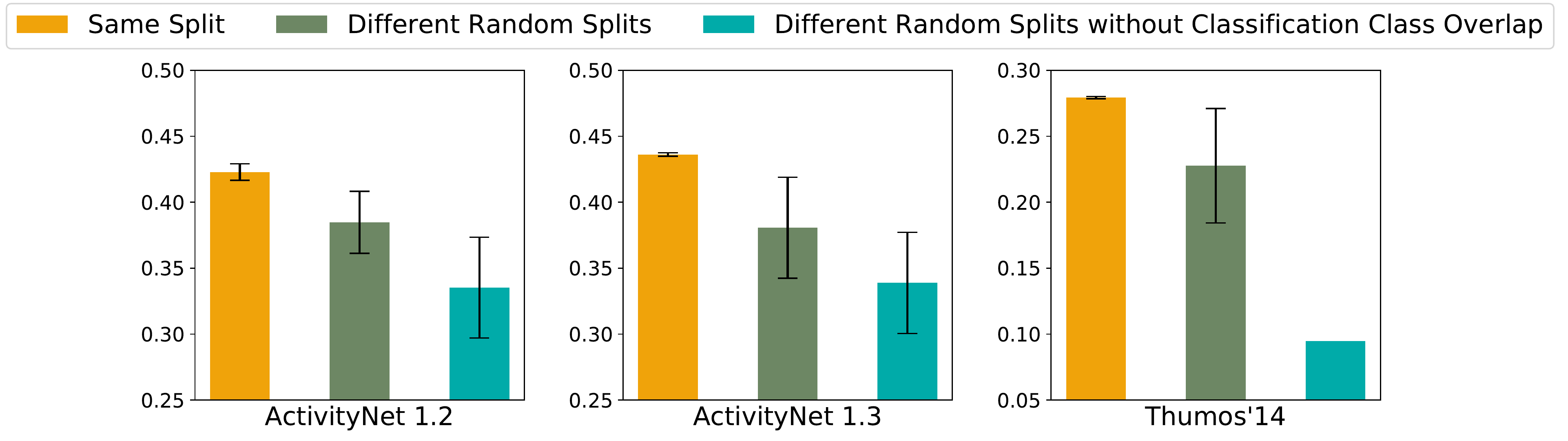} \\ ~ \\
    \caption{Mean and standard deviation of mAP@0.5 results on different base and novel class splits of ActivityNet1.2, ActivityNet1.3 and THUMOS'14 for our one-shot F-PAD model. }
    \label{fig:results_dataSplit}
    %\vspace{10pt}
\end{figure*}

We run our one-shot model on the six different base and novel splits of each data set described in the experimental setup section~\ref{exp:different_setup}.
The mean and standard deviation of the mAP@0.5 results are shown in Figure~\ref{fig:results_dataSplit}.
The yellow histogram shows the mean and standard deviation of three different runs on ``randomSplit1".
From the yellow histogram, we can see that though there is certain randomness of sampling one shot example in each iteration, on the same data split, the mAP@.5 results of three different runs are quite stable with small standard deviation.
The green histogram shows the mean and standard deviation of three runs on three different random base and novel splits without control on class overlap, namely ``randomSplit1", ``randomSplit2" and ``randomSplit3".
The cyan histogram shows the mean and standard deviation of three runs on three random base and novel splits with no overlap between novel classes and classes in Sports-1M.
The corresponding data-splits are  ``controlled\_randomSplit1", ``controlled\_randomSplit2", ``controlled\_randomSplit3".
Comparing the green histogram with the cyan histogram, we observe that the mean mAP@0.5 decreases as novel classes become completely disjoint from pretraining classes.
The standard deviations on both sets of splits (green histogram and cyan histogram) are relatively large.
The reason might be that the three random splits might have very different demography between the novel and the base classes.
For example, one split could have all the novel classes belonging to aquatic sports while none is present in the base class while another split may have the presence of aquatic sports (belonging to disjoint classes) present in both base and novel classes.
We argue that splitting randomly and showing the standard deviation on the performance would help the community to understand the robustness of the methods better.
Note that for Thumos'14, there is only one controlled split without class overlap, thus no standard deviation is shown for it, and also the number of base and novel classes in this split is different on this dataset (cyan histogram vs green histogram) and thus results can't be directly compared.

\subsection{One-shot and Five-shot Result Comparison}
\label{exp:oneShot_and_fiveShot}

In Table~\ref{tab:res_final}, we choose two base and novel splits ``randomSplit1" and ``\seqsplit{controlled\_randomSplit1}" from each dataset and report the one-shot and five-shot results of our \modelname~in terms of mAP@0.5 and average mAP to show the relative improvement of increasing the number of shots. 
Though we list the results from existing work~\cite{yang2018one}, based on our observation in previous section that the standard deviation is high between different base and novel splits, these results are not directly comparable to us since neither the split nor the code is public. 
In general, the performance of our five-shot model (F-PAD@5) is better than the one-shot model (F-PAD@1) significantly which shows the benefit of our end-to-end model in a scenario where a few more example is available for training.
Besides, we find that, on  ActivityNet1.2 and  ActivityNet1.3, the relative improvement is less prominent for the controlled split (last two rows) compared to the uncontrolled split ($4^{th}$ and $3^{rd}$ last rows) when going from 1-shot to 5-shot scenario.
For ``randomSplit1" and ``controlled\_randomSplit1" on Thumos14 dataset, the number of base and novel classes in these two splits  are different, thus results can't be directly compared.
``randomSplit1" has 6 classes for training and 14 classes for testing following~\cite{yang2018one}, while ``controlled\_randomSplit1" has 11 classes for training and 9 classes for testing since we can only find 9 unpolluted classes for novel classes.
However, we can observe that there is a significant drop in performance in ``controlled\_randomSplit1" compared to ``randomSplit1".
This might be because in ``randomSplit1" certain novel classes overlapping with the pretraining classes have high per-class AP and thus bringing the overall mean AP up.

\begin{table}[!t]
\centering 
\caption{
%Activity detection results on activityNet1.2
Few-shot temporal activity detection results on ActivityNet1.2, ActivityNet1.3 and THUMOS'14 (in percentage). For  ActivityNet1.2 and ActivityNet1.3, we report mAP at tIoU threshold $\alpha=0.5$ and average mAP of $\alpha \in \{0.5,0.95\}$. For  THUMOS'14, we report mAP at tIoU threshold $\alpha=0.5$. @1 means ``one-shot" and @5 means ``five-shot". Note that the results from paper~\cite{yang2018one} are not directly comparable to us since the base and novel split is unknown according to the analysis in Sec.~\ref{exp:different_split}
%The results with $^*$ use the activityNet1.3 pretrained two-stage proposal weights as initialization.
}
\label{tab:res_final}
\resizebox{0.9\linewidth}{!}{
\begin{tabular}{l | c | c c | c c| c } 
\Xhline{3\arrayrulewidth} 
&  \multirow{2}{*}{base/novel split} & \multicolumn{2}{c|}{ActivityNet 1.2} & \multicolumn{2}{c|}{ActivityNet 1.3} & THUMOS’14 \\
 \cline{3-7}
 & &  \makecell{mAP@0.5} & \makecell{average \\ mAP} & \makecell{mAP@0.5} & \makecell{average \\ mAP} & \makecell{mAP@0.5} \\
\Xhline{3\arrayrulewidth} 
 CDC@1~\cite{yang2018one} &  unknown & 8.2 & 2.4 & - & - & 6.4\\ %\hline
CDC@5~\cite{yang2018one} & unknown & 8.6 & 2.5 & - & - & 6.5\\ %\hline
 sliding window@1~\cite{yang2018one} &  unknown &  22.3 & 9.8 & - & - & 13.6 \\ %\hline
 sliding window@5~\cite{yang2018one} &  unknown & 23.1 & 10.0 & - & - & 14.0 \\ \hline
%   F-PAD@1  &  0.4154954015  & 0.2851123363  \\ %\hline
%  F-PAD@5  &  0.5082145832 & 0.3421236028  \\ \hline
 F-PAD@1  & randomSplit1 & 41.5 & 28.5 & 43.4 & 29.1 & 24.8 \\ %\hline
 F-PAD@5  & randomSplit1 & 50.8 & 34.2 &  51.6 & 34.6 & 28.1\\ %\hline
  F-PAD@1  & controlled\_randomSplit1 & 31.7 & 19.4 & 31.4 & 20.8 & 9.5 \\ %\hline
 F-PAD@5  & controlled\_randomSplit1 & 37.9 & 23.5 &  39.0 & 24.1 & 10.3\\ \hline

  \end{tabular}}
%\vspace{5pt}
\end{table}

%\vspace{10pt}
\section{Conclusion}

In this paper, we propose a few-shot temporal activity detection model based on proposal regression, which is composed of a temporal proposal subnet and a few-shot classification branch based on similarity values.
Our model is end-to-end trainable and can benefit from observing additional few-shot examples.
Moreover we take the frame rate differences between the few-shot video inputs and temporal proposals into consideration and apply one distributional domain adaptation loss to minimize the feature differences brought by the frame rate differences.
Besides, we also observe that the few-shot activity detection performance on novel classes could be affected by the class similarity between novel classes and pretrained classification classes, as well as base classes.
Suggesting and coming up with fairer evaluation strategy for few-shot activity detection taking the class similarity into consideration will be a future work for this.

\clearpage

\bibliographystyle{splncs04}
\bibliography{egbib}
\end{document}